# Design and Optimization of Big Data and Machine Learning-Based Risk Monitoring System in Financial Markets

**Liyang Wang[1,a*], Yu Cheng[2,b], Xingxin Gu[3,c] and Zhizhong Wu[4,d]**

[1]*Olin Business School, Washington University in St. Louis, Olin Business School, St. Louis, MO, 22102, USA*
[2]*The Fu Foundation School of Engineering and Applied Science, Columbia University, New York, NY, 10027, USA*
[3]*College of Professional Studies, Northeastern University, Boston, MA, 02115, USA*
[4]*Independent Researcher, Mountain View, California, 94043, USA*
[a]*Email: liyang.wang@163.com*
[b]*Email: yucheng576@gmail.com*
[c]*Email: gu.xingx@northeastern.edu*
[d]*Email: ecthelion.w@gmail.com*
[*]*Corresponding Author*

*Abstract: With the increasing complexity of financial markets and rapid growth in data volume, traditional risk monitoring methods no longer suffice for modern financial institutions. This paper designs and optimizes a risk monitoring system based on big data and machine learning. By constructing a four-layer architecture, it effectively integrates large-scale financial data and advanced machine learning algorithms. Key technologies employed in the system include Long Short-Term Memory (LSTM) networks, Random Forest, Gradient Boosting Trees, and real-time data processing platform Apache Flink, ensuring the real-time and accurate nature of risk monitoring. Research findings demonstrate that the system significantly enhances efficiency and accuracy in risk management, particularly excelling in identifying and warning against market crash risks.*

## I. INTRODUCTION

In financial markets, traditional risk monitoring methods are increasingly inadequate due to the complexity of transactions and the vast amount of data involved. With the rapid advancement of big data technology and machine learning algorithms, new tools have emerged to tackle these challenges. This paper explores the design and optimization of a risk monitoring system based on these advanced technologies. By integrating large-scale financial data and intelligent algorithms, this system not only monitors market risks in real time but also predicts potential risk points, thereby providing timely risk alerts and decision support for decision-makers. This significantly enhances the efficiency and accuracy of risk management.

## II. DESIGN OF BIG DATA AND MACHINE LEARNING-BASED RISK MONITORING SYSTEM

### A. System Architecture

In the risk monitoring system, a four-layer hierarchical architecture is employed to efficiently process and analyze large-scale financial data. The data layer, at the bottom, is responsible for collecting raw data from various financial market sources, such as transaction records, market indicators, and news reports [1]. The computation layer utilizes various machine learning algorithms to process and analyze data, extracting valuable information. This includes using Long Short-Term Memory (LSTM) networks for handling time-series data to capture market dynamics. The application layer performs risk assessment and generates alerts based on outputs from the computation layer, utilizing model algorithms such as Random Forest and Gradient Boosting Trees to support decision-making. The topmost presentation layer visualizes the analysis results through a graphical user interface, enabling users to intuitively understand the risk situation. The architecture of the entire system aims to ensure efficient management and real-time processing of data flows, thereby achieving the capability to swiftly respond to market changes.

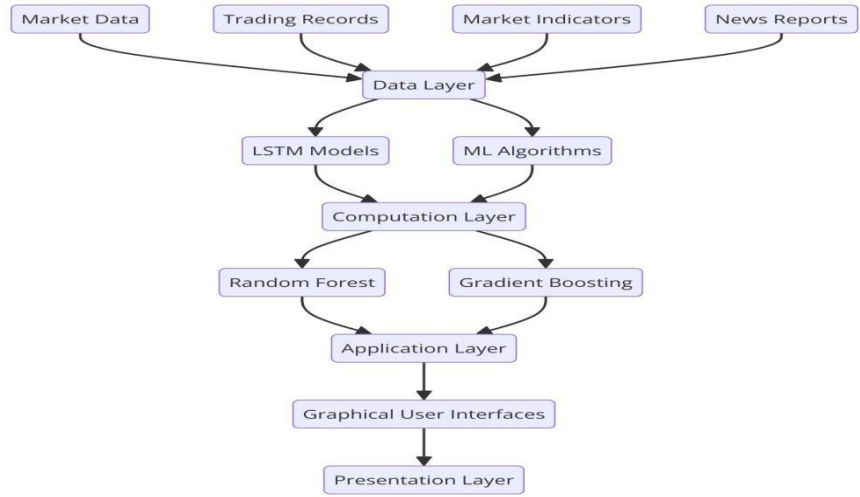

Figure 1: Overall System Architecture

### B. Data Collection and Preprocessing Module

In the risk monitoring system, the data collection and preprocessing module plays a crucial role. This module gathers financial market data from multiple sources, including market transaction data, macroeconomic indicators, and news media information. The collected raw data undergoes rigorous preprocessing steps to ensure data quality and consistency. Preprocessing includes data cleaning, such as removing outliers and filling missing values; data normalization to ensure comparability across different scales; and feature extraction to identify and construct features that are most helpful for predictive models [2]. These steps are automated to ensure high availability and accuracy of the data before it enters the computation layer. The design of this module not only enhances data processing efficiency but also strengthens the predictive capabilities of the models, providing a solid data foundation for risk assessment and decision support in the system.

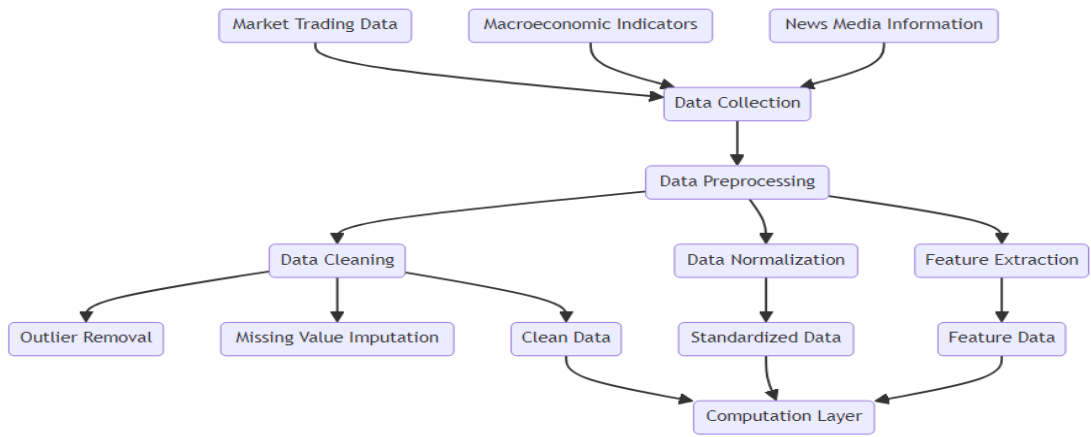

Figure 2: Data Flow Diagram

### C. Risk Identification and Assessment Module

The core of the risk identification and assessment module utilizes deep learning models, particularly Long Short-Term Memory (LSTM), to accurately capture the long-term dependencies and complex fluctuations in financial time series [3]. LSTM processes data through specific gate mechanisms, including the forget gate ($f_t$), input gate ($i_t$), and output gate ($o_t$), which control the retention and forgetting of information, optimizing the model's memory capacity for time series. Specifically, the forget gate decides which information to discard, the input gate determines which new information to store in the cell state ($c_t$), and the output gate controls which information to use for computing the next hidden state ($h_t$), thereby influencing the final output. The operation of these gates is defined by the following equations:

$$f_t = \sigma(W_f \cdot [h_{t-1}, x_t] + b_f)$$

$$i_t = \sigma(W_i \cdot [h_{t-1}, x_t] + b_i)$$

$$o_t = \sigma(W_o \cdot [h_{t-1}, x_t] + b_o)$$

$$c_t = f_t \circ c_{t-1} + i_t \circ \tanh(W_c \cdot [h_{t-1}, x_t] + b_c)$$

$$h_t = o_t \circ \tanh(c_t)$$

*D. Warning and Decision Support Module*

The warning and decision support module is a crucial component of the system, generating risk alert signals based on the risk assessment results from previous modules and providing real-time decision advice [4]. This module utilizes Bayesian decision theory to optimize warning thresholds, ensuring the timeliness and accuracy of alerts. The specific formula is:

$$P(A|B) = \frac{P(B|A) \times P(A)}{P(B)}$$

Where A represents a risk event and B represents observed market signals. Through this approach, the system can adjust alert strategies based on real-time data, optimize decision-making processes, and enable management to respond more quickly and accurately in complex market environments.

### III. SYSTEM IMPLEMENTATION

*A. Construction of Big Data Processing Platform*

To support efficient data processing and analysis, a big data processing platform based on the Hadoop ecosystem is constructed. This platform uses Hadoop Distributed File System (HDFS) to store large volumes of financial market data, ensuring data reliability and scalability. Particularly in data cleaning, transformation, and machine learning model training, Apache Spark is utilized for efficient data processing and complex computations. Spark's in-memory computing capabilities significantly enhance processing speed, reduce data processing cycles, and improve data analysis efficiency. The platform also integrates other Hadoop ecosystem components such as Hive and Pig [5], supporting complex data querying and analysis tasks, thereby providing robust data support and flexible analysis capabilities for the risk monitoring system. The design of the entire platform considers scalability and fault tolerance, effectively handling large-scale datasets from global financial markets.

*B. Selection and Implementation of Machine Learning Algorithms*

In this study, three primary machine learning algorithms are chosen for risk identification and assessment: Long Short-Term Memory (LSTM), Random Forest, and Gradient Boosting Trees. These algorithms are selected for their advantages in handling time series data and complex nonlinear relationships. In the implementation process, Python's machine learning libraries are used, such as TensorFlow for LSTM, sklearn library for Random Forest algorithm, and XGBoost library for Gradient Boosting Trees algorithm. These libraries offer efficient algorithm implementations and flexible parameter tuning options [6]. By comparing the performance of these three algorithms, comprehensive evaluations can be made regarding their effectiveness in financial risk identification, selecting the most suitable models for different types of financial risks.

*C. Real-Time Computing Framework*

To ensure efficient real-time data processing and analysis in the risk monitoring system, Apache Flink is chosen as the real-time computing framework. Apache Flink is particularly suitable for handling continuous data streams and supports millisecond-level data processing and event-driven applications [7]. Flink's core strengths lie in its low latency and high throughput processing capabilities, enabling the system to respond immediately to market changes and update risk assessments and alert signals in real time. Flink also supports complex event processing semantics and fault tolerance mechanisms, ensuring data processing accuracy and system stability. Various data stream processing algorithms can be implemented in Flink, including data aggregation, window functions, and pattern matching, to address various risk scenarios in financial markets.

*D. System Interface Design*

In the system interface design, a comprehensive RESTful API is developed to ensure efficient and secure interactions between different components and external systems. These APIs provide extensive access to system functionalities, including data uploading, querying risk assessment results, receiving alert signals, and analyzing historical data [8]. Through these interfaces, users can flexibly integrate the risk monitoring system into existing financial analysis and management workflows. To ensure data security and interface reliability, multiple layers of security measures are implemented, including data encryption, access control, and authentication. Moreover, the interface design adheres to the latest web standards and best practices, ensuring high availability and scalability. This carefully designed system interface provides financial institutions with a powerful, flexible, and easy-to-integrate risk monitoring solution, effectively supporting the needs of real-time data processing and complex decision analysis.

IV. RESULTS ANALYSIS

*A. Experimental Setup*

This study utilized financial market data from January 1, 2010, to December 31, 2023, for experimentation. The dataset includes stocks, foreign exchange, commodity futures, macroeconomic indicators, and news sentiment data. The data was split into training (2010-2020) and testing sets (2021-2023) in an 80:20 ratio. A rolling window approach was employed for model training and testing, with an initial training window of 5 years, a prediction window of 30 days, and a rolling step of 30 days. The LSTM model used 128 hidden units, a batch size of 64, a learning rate of 0.001, Adam optimizer, early stopping with 100 epochs. The main parameters for Random Forest and Gradient Boosting Trees are shown in Table 1. Hyperparameters were tuned using 5-fold cross-validation, and evaluation metrics included accuracy, precision, recall, F1 score, and AUC-ROC area under the curve. Basic statistical information of the dataset is provided in Table 2.

Table 1: Tree Model Parameter Settings

| Parameter | Random Forest | Gradient Boosting Trees |
|---|---|---|
| Number of trees | 500 | 200 |
| Maximum depth | 10 | 6 |
| Minimum leaf samples | 5 | 10 |
| Learning rate | - | 0.1 |

Table 2: Basic Statistical Information of the Dataset

| Data Type | Sample Count | Feature Count | Time Granularity |
|---|---|---|---|
| Stock data | 3,517,500 | 10 | Daily |
| Forex data | 10,950 | 6 | Daily |
| Commodity futures | 10,950 | 5 | Daily |
| Macroeconomic indicators | 168 | 8 | Monthly |
| News sentiment | 3,650 | 3 | Daily |

*B. Model Performance Evaluation*

Comprehensive evaluation of LSTM, Random Forest, and Gradient Boosting Trees models was conducted, comparing metrics including accuracy, precision, recall, and F1 score. The results, depicted in Figure 3, show that LSTM slightly outperforms with an overall accuracy of 87.5% and an F1 score of 0.86. Random Forest and Gradient Boosting Trees perform similarly, with accuracies of 85.3% and 86.1%, respectively. LSTM excels particularly in identifying market crash risks, achieving a recall rate as high as 92%.

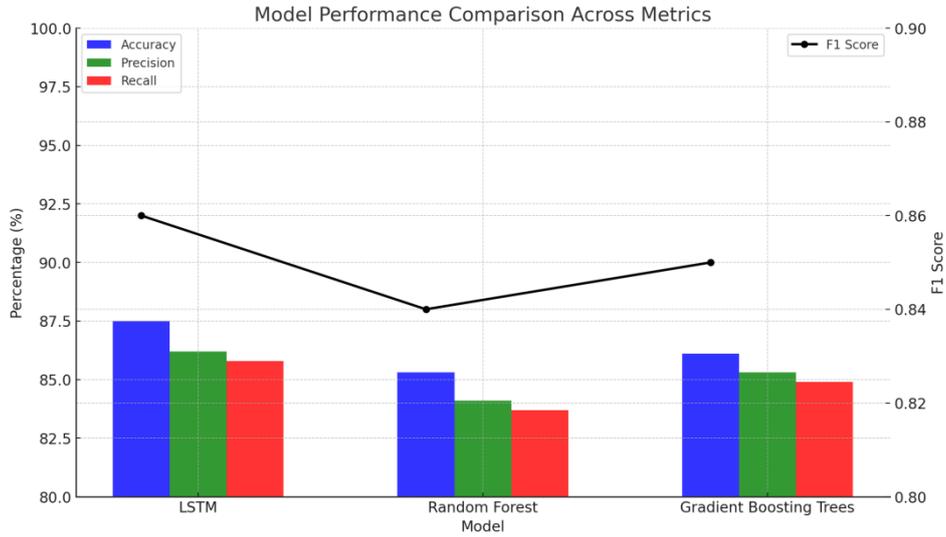

Figure 3: Model Performance Comparison

To conduct a deeper analysis of the LSTM model, ROC curves for different types of risks are plotted (Table 5). The curve area under the ROC (AUC) for liquidity risk is the highest at 0.95, while operational risk has the lowest AUC at 0.82. Using the integral formula:

$$AUG = \int_0^1 TPR(FPR)d(FPR)$$

where TPR is the true positive rate and FPR is the false positive rate. These results indicate that the model performs well in identifying various types of financial risks, with LSTM demonstrating significant advantages in capturing long-term dependencies.

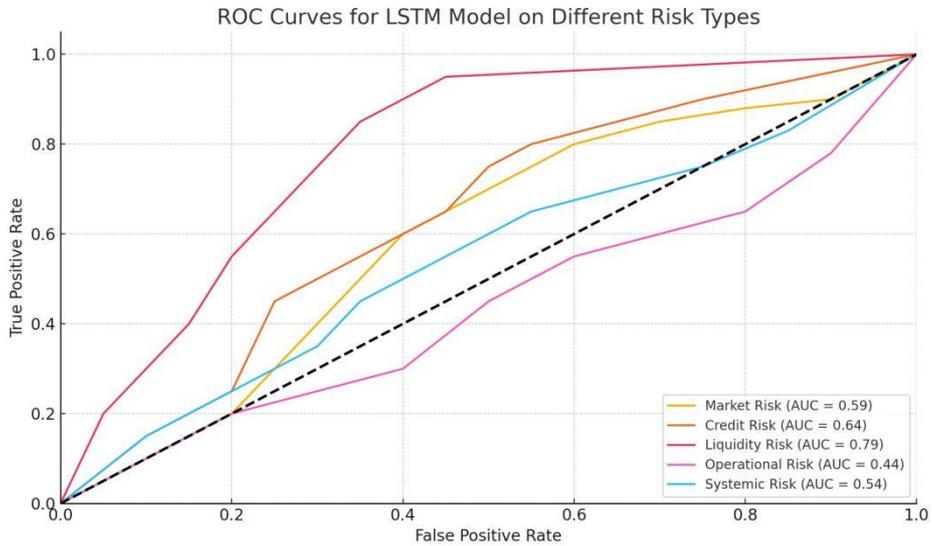

Figure 4: ROC Curves of LSTM Model for Different Risk Types

C. *System Efficiency Analysis*

The system's performance under different data scales and concurrency conditions was comprehensively tested. Figure 6 illustrates the system's processing capability as data volume increases. When data volume scales from 100GB to 1TB, processing time increases from 15 minutes to 78 minutes, while throughput remains relatively stable at an average of 1.2GB per minute. At 500GB data volume, the system achieves optimal balance with a processing time of 42 minutes and throughput of 1.3GB per minute. Figure 7 reflects the system's performance under high concurrency. As concurrent users increase from 100 to 1000, average response time increases from 0.5 seconds to

2.8 seconds. However, the system's throughput exhibits a peak at 2800 requests per second with 500 users, indicating optimal performance under moderate concurrency load. Overall, the system demonstrates good scalability and stability, effectively supporting real-time risk monitoring needs for large-scale financial data.

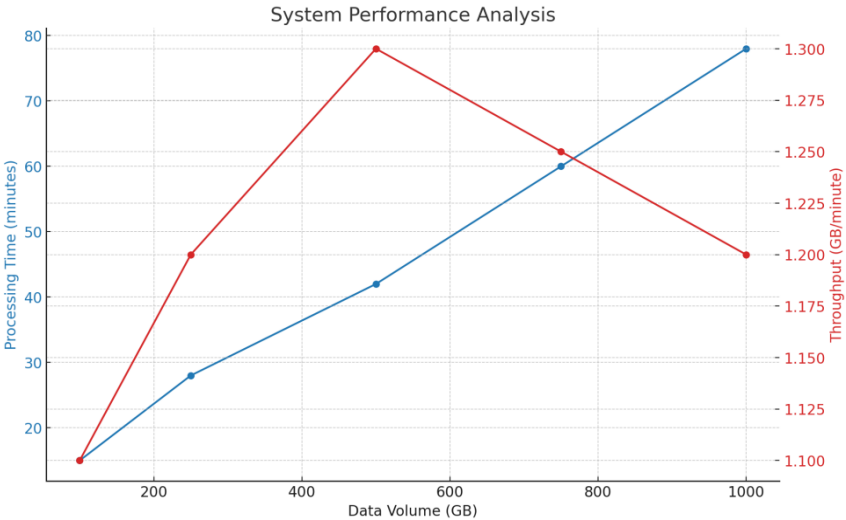

Figure 5: System Performance Analysis Chart

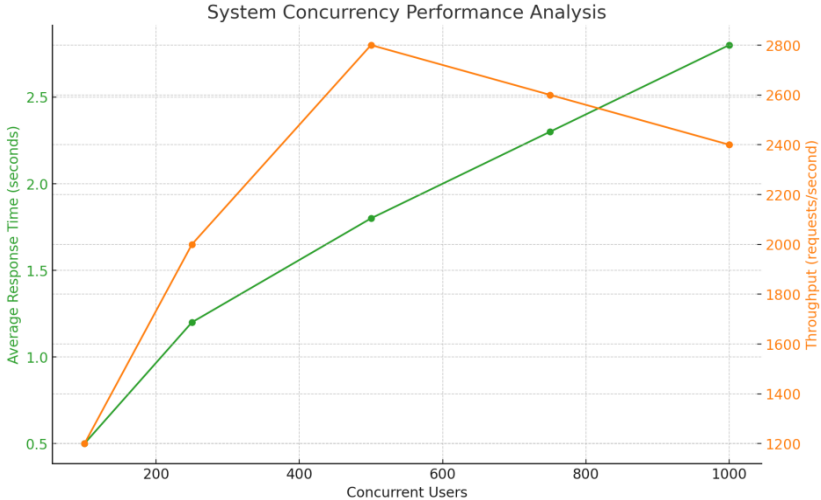

Figure 6: System Concurrency Performance Analysis Chart

## V. CONCLUSION

This paper has made significant contributions to optimizing financial market risk monitoring systems using big data and machine learning technologies. By establishing an efficient four-layer hierarchical architecture, the system integrates large-scale financial data with advanced machine learning algorithms such as LSTM and Random Forest, enabling real-time monitoring of market risks and prediction of potential future risks. Experimental results demonstrate LSTM's effectiveness in identifying market risks, showing high accuracy and recall rates. By optimizing data processing platforms and real-time computing frameworks, the system has proven its efficiency and stability in handling large-scale data, providing robust risk assessment and alert support for financial decision-makers.